# Towards A Structured Overview of Use Cases for Natural Language Processing in the Legal Domain: A German Perspective

*Completed Research Full Paper*

**Juraj Vladika**
Technical University of Munich
juraj.vladika@tum.de

**Stephen Meisenbacher**
Technical University of Munich
stephen.meisenbacher@tum.de

**Martina Preis**
Technical University of Munich
martina.preis@tum.de

**Alexandra Klymenko**
Technical University of Munich
alexandra.klymenko@tum.de

**Florian Matthes**
Technical University of Munich
matthes@tum.de

## Abstract

In recent years, the field of Legal Tech has risen in prevalence, as the Natural Language Processing (NLP) and legal disciplines have combined forces to digitalize legal processes. Amidst the steady flow of research solutions stemming from the NLP domain, the study of use cases has fallen behind, leading to a number of innovative technical methods without a place in practice. In this work, we aim to build a structured overview of Legal Tech use cases, grounded in NLP literature, but also supplemented by voices from legal practice in Germany. Based upon a Systematic Literature Review, we identify seven categories of NLP technologies for the legal domain, which are then studied in juxtaposition to 22 legal use cases. In the investigation of these use cases, we identify 15 ethical, legal, and social aspects (ELSA), shedding light on the potential concerns of digitally transforming the legal domain.

## Introduction

The world of law is characterized by vast amounts of textual data in the form of documents like statutes, contracts, legislature, court rulings, and legal opinions. The daily work of legal professionals often includes tedious reading tasks and manual analysis of legal documents. In the digital era, marked by information overload and huge amounts of new data produced every day, the ability to extract insights quickly and accurately from large legal repositories has become even more important. The ongoing digital transformation and advancements in artificial intelligence (AI) are affecting numerous fields, among them the legal realm. Considering the textual nature of legal resources, an especially important field in helping to automate repetitive legal tasks is Natural Language Processing (NLP). Advances in NLP, underpinning the recent AI boom, also make a clear and promising case for its use in the legal domain (Zhong et al. 2020).

Applying technical solutions such as AI to the legal domain is commonly known as Legal Tech. NLP solutions can be used to improve access to justice, reduce costs, and increase efficiency. The adoption of NLP in the legal sector prompts a new set of pressing ethical questions, such as algorithmic discrimination, inclusion of minority languages, transparency, trust, and the equality of capabilities between companies and public institutions (Hildebrandt 2021; Djeffal and Horst 2021).

While a considerable amount of research has been conducted in Legal Tech regarding the development of new tools to facilitate the automation of legal services, there have been no studies that provide a systematic overview of the use cases of NLP in the legal sector. Most importantly, the opinion of legal professionals on the use and adoption of NLP tools in their daily work has yet to be surveyed. This includes addressing the ethical, legal, and social aspects (ELSA) arising when such technologies are adopted (Zwart et al. 2014).

To bridge these research gaps, our study aims to provide a structured overview of NLP use cases in the legal domain and a more in-depth view of their ethical, legal, and social implications.

To accomplish the goals of our work, the following research question has been defined:

**RQ**: What are the legal use cases in which NLP technologies can be utilized?

In order to answer the above research question, we follow a two-part research methodology, firstly conducting a Systematic Literature Review surveying the predominant NLP technologies used for the legal domain. Then, we conduct a series of Semi-Structured Interviews with legal professionals in Germany, hoping to gain a practical perspective on the use cases identified from the literature.

In conducting this research, we make the following contributions to the current body of knowledge:

1. To the best of our knowledge, we provide the first structured overview of use cases for NLP in the legal domain.
2. We investigate the ethical, legal, and social aspects of NLP in the legal domain.
3. We ground all of our findings in interviews conducted with legal professionals working in Germany, adding a practical perspective to the above findings.

# Theoretical Background

## Legal Tech and Legal NLP

Legal Technology, often called "Legal Tech", is a term that broadly refers to the innovative technology and software to streamline and enhance legal services (Corrales et al. 2019). It covers all information technologies used in the legal service sector such as contract or document management systems, systems of e-discovery in litigation, or judicial predictive systems (Mania 2022). The motivation behind deploying Legal Tech is that with the help of technology, legal work and access to justice can be improved, becoming cheaper and more accessible for a greater proportion of the population (Hartung et al. 2017).

Natural Language Processing (NLP) can be defined as an area of research and application that explores how computers can be used to understand and manipulate natural language text or speech to do useful things (Chowdhary 2020). The most popular discipline within Legal Tech has since its inception been NLP, considering the dependence of law on written resources (Dale 2018), thus leading to the notion of "Legal NLP". Zhong et al. (2020) provide an overview of legal AI, concluding that legal professionals often think about solving tasks with rule-based and symbolic methods, while NLP researchers concentrate more on data-driven and embedding methods or methods based on numeric representations of text.

The gap between legal NLP research and practitioners' needs and use of this technology (Zhong et al. 2020) serves as a motivation for this study. To the best of our knowledge, this is the first systematic review of NLP in the legal domain, particularly one augmented by semi-structured interviews with legal professionals. In this work, we specifically focus on NLP for Legal Tech and use Legal Tech interchangeably with Legal NLP.

## Ethical, Legal, Social Aspects (ELSA)

The term "Ethical, Legal and Social Aspects", often abbreviated as ELSA, represents a framework used to analyze the implications of emerging technologies on individuals and society (Zwart et al. 2014). Since its inception in 1988, the framework has been used for numerous other fields and emerging technologies, among them AI (Čartolovni et al. 2022). By systematically addressing ethical, legal, and social dimensions, stakeholders can work to mitigate potential risks and challenges, foster public trust, and promote the responsible and equitable advancement of technology (Carter et al. 2020). This especially applies to NLP and AI in the legal domain, considering that law inherently involves humans and affects countless individuals and society in general (McPeak 2018).

# Methodology

The research design is structured as a hybrid, two-part study, in which a Systematic Literature Review (SLR) is performed. The findings of the literature review are complemented with semi-structured interviews (SSIs) with legal domain experts. This serves not only to validate themes and constructs identified in the literature review process, but also to gain new insights which may not be reflected in the selected literature, especially concerning the practical needs of legal professionals in their daily work.

## *Systematic Literature Review*

The SLR conducted follows the methodology introduced by Kitchenham et al. (2009). The study begins by defining search strings related to Legal Tech and NLP which are used to query well-known academic databases of technical publications, namely IEEE Xplore, ACM Digital Library, and Scopus, as well as the Association of Computational Linguistics (ACL) Anthology, the predominant digital library for publications in computational linguistics and NLP. For the SLR, we do not consult legal sources, as the SLR (and our work) is grounded in NLP technologies, from which use cases can be extracted.

Leveraging the abovementioned electronic sources, we employ the following search string:

("NLP" AND "Legal") OR ("NLP" AND "Law") OR ("Natural Language Processing" AND "Law") OR ("Natural Language Processing" AND "Legal ") OR ("LegalTech" AND "Use Case") OR ("Legal Tech" AND "Use Case")

Note that due to limitations of the ACL Anthology, we used a simple search with the keywords *legal* and *law* (as the database is already NLP-centric). Altogether, these searches resulted in 122 results.

To aid in the filtering process, we first apply exclusion criteria, namely (1) not openly or institutionally accessible, or (2) not written in English. This narrowed down the results to 119 papers. Next, paper abstracts were screened according to further exclusion criteria, namely: (1) no legal relation − 45 papers, (2) no discernable legal use case − 13 papers, (3) not a peer-reviewed publication − 3 papers, (4) poor quality − 4 papers, and (5) duplicates − 1 paper. This resulted in a final set of **49 papers**.

With these papers, the first step involved a team of researchers reading the papers for comprehension. Next, each paper was analyzed to extract four points of information: (1) publication year, (2) relevant NLP technology, (3) ELSA, (4) legal use case(s). The ultimate goal of this process was to construct overall trends in the Legal Tech research landscape, particularly regarding the prevalence of NLP technologies for the legal domain, as well as a quantification of use cases proposed in the selected works.

## *Semi-Structured Interviews*

The second step consisted of SSIs (Adams 2015) with domain experts. The goal of these interviews was to gain additional findings that serve to augment the findings of the SLR, in a way that facilitates open discussions leading to insights on the use cases of NLP in the legal domain.

The instrumentation to be utilized for the interview study consists of a pre-defined interview guide, which was provided to interviewees prior to the conducted interview. The questionnaire consists of questions related to the practical needs of users of legal tech solutions, obstacles to successful adoption, and insights into currently relevant issues. The semi-structured nature of the interviews allowed for follow-up questions, and clarifications as deemed necessary by the interviewer. This enables the interviewer to focus on new insights, which may not have been gained from the SLR alone. Interviews were conducted solely online, and the study continued until a saturation of themes could be observed, in line with constant comparison.

To construct the interview guide, we followed the methodology of Kallio et al. (2016). In particular, we used our retrieved knowledge from the SLR as the basis of the guide, and collectively formulated questions that were intentionally open-ended (McIntosh and Morse 2015). The guide was internally pilot-tested on two colleagues, and subsequently field-tested on one interviewee (I1), which revealed that no updates to the question were needed. Each interview began with an overview of our study objective and obtaining consent to record the interview. The remaining interview guide consisted of five questions:

- *Do you use digital aids to complete your daily tasks? Which ones?*
- *Which repetitive tasks are part of your everyday work?*
- *Where do you see future use cases for legal tech?*
- *Where do you think legal tech will develop in the next few years?*

- *Do you have ethical, legal, or social concerns about the use of technology in the legal sector? If so, what are they?*

Following each interview, the results were transcribed and analyzed, following a thematic content analysis (Braun and Clarke 2006). The main goal of this analysis is to build up themes and constructs from the interview data, which can be used to produce generalizable findings from the interview study. The process was followed by a team of three researchers and consisted of five steps: (1) review of transcripts, (2) initially coding the transcripts to highlight important segments, (3) conceptualizing themes based on the codes, (4) reviewing the themes, and (5) naming the themes. As this analysis was performed in an iterative fashion, identified themes could be merged, refuted, or otherwise updated as new interview data was considered. In the end, the purpose of performing the content analysis was to augment the structured findings from the SLR, as well as to determine a saturation point, i.e., when no new themes were discovered after an interview.

The list of 18 interview study participants is shown in Table 1. The interviewees consented to the findings being published in an anonymized manner but did not agree to disclose the full transcriptions. The findings from the SSIs are referenced in the following by their identifying code, as found in Table 1. Note that all interviewees work in Germany, so the region of each interviewee is not tabulated explicitly.

| Code | Position | Organization | Experience (years) | Duration (min) |
|------|----------|--------------|--------------------|----------------|
| I-1 | Researcher and Attorney | Large-sized | 9 | 31 |
| I-2 | Attorney | Medium-sized | 11 | 58 |
| I-3 | Attorney | Large-sized | 9 | 31 |
| I-4 | Law Student | Student | 5 | 35 |
| I-5 | Attorney | Large-sized | 29 | 19 |
| I-6 | Judge in Ministry of Justice | State Institution | 16 | 61 |
| I-7 | Law Student | Student | 6 | 38 |
| I-8 | Attorney | Micro-sized | 14 | 55 |
| I-9 | Attorney | Medium-sized | 27 | 55 |
| I-10 | Notary | Micro-sized | 9 | 36 |
| I-11 | Notary | Small-sized | 29 | 42 |
| I-12 | Attorney | Small-sized | 12 | 30 |
| I-13 | Judge | State Institution | 10 | 58 |
| I-14 | Researcher and Attorney | Large-sized | 7 | 36 |
| I-15 | Researcher and Attorney | Large-sized | 10 | 23 |
| I-16 | Attorney | Large-sized | 9 | 36 |
| I-17 | Judge | State Institution | 13 | 47 |
| I-18 | Judge | State Institution | 18 | 49 |
| | **Average:** | | **12.3** | **43.5** |

**Table 1. Interview Study Participants**

# Results

The insights discovered during the SLR and SSIs were combined to garner a holistic picture of the current trends in the legal NLP research landscape, including NLP technologies used, legal practitioners' use cases and needs, and practical obstacles to adoption in the form of ELSA concerns.

## *NLP Technologies for the Legal Domain*

| NLP Category | Description | # |
|--------------|-------------|---|
| Information Extraction | Extracting relevant entities from text, such as legal actors, legal matters, witnesses, cases. | 18 |
| Text Classification | Categorizing written text into predefined categories, such as deciding on the verdict of a court ruling or on the legality of a contract. | 17 |
| Document Analysis | Processing of legal documents, like contracts or court rulings. Includes NLP tasks such as sentiment analysis, readability assessment, etc. | 12 |
| Text Representation | Training of deep learning models to represent legal text efficiently, usually with language models. | 9 |
| Text Generation | Using generative NLP models to produce new text, for use cases like translation, summarization, etc. | 5 |
| Conversational NLP | Encompasses question answering and any work with conversational agents assisting in legal work. | 4 |
| Overview Papers | Surveys, discussions, perspective papers, etc. | 4 |
| Syntactic Analysis | Linguistic assessment of written text by breaking down sentences into its syntactic components. | 3 |

**Table 2. Categories of NLP technologies found in the SLR, with representative sources**

Table 2 shows the results of the SLR by tabulating the identified NLP technology, count of appearances from the SLR, and short description. The papers were categorized into eight categories, guided by our expertise from working in NLP and Legal Tech. The assignment of technical papers into categories was performed by three researchers, two of whom hold over five years of research experience in NLP.

The most frequent category was information extraction, which includes detecting and extracting the most important entities, relations, or events from legal text. Examples include modificatory provisions (Mazzei et al. 2009), reference extraction (Bach et al. 2019), and identifying legal party members (de Almeida et al. 2021). Another category is text classification, which involves classifying text into predefined categories. Examples are topic discovery (Vianna and de Moura 2022; Braun and Matthes 2022), GDPR compliance check (Hamdani et al. 2021), and legal judgment prediction (Semo et al. 2022).

Document analysis includes all tasks involving the processing of legal documents. Examples include patent retrieval (Andersson et al. 2016), litigation analytics (Vacek et al. 2019), and document segmentation (Seyler et al. 2020). Text representation deals with representing legal text in high-dimensional vector spaces, fit for deep learning models. This includes the development of transformer-based NLP models for the legal domain, such as LegalBERT (Mamakas, et al. 2022).

Text generation includes tasks in producing text, with the most prominent task being text summarization (Merchant and Pande 2018; Glaser et al. 2021). Conversational NLP focuses on providing answers to legal questions (McElvain et al. 2019; Sovrano et al. 2020) and the development of dialogue systems that can assist legal professionals or clients with legal matters (Shubhashri et al. 2018; Hong et al. 2021). Syntactic analysis involves processing text on a syntactic level, by a linguistic analysis (Garofalakis et al. 2016; Saxena et al. 2021). The search also included two surveys (Krasadakis et al. 2021; Conrad et al. 2023).

## *Legal Use Cases of NLP*

| Category | Use Cases | Description | SLR | SSI |
|---|---|---|---|---|
| Trustworthiness | Automation of Auditing | Streamlining the auditing of legal documents, contracts, financial errors. | 1 | 0 |
| | GDPR Compliance Check | Automatic checking of whether a document complies with the General Data Protection Regulation (GDPR). | 2 | 0 |
| | Risk Assessment | Evaluation of risks associated with legal matters or business activities. | 1 | 1 |
| Document Analysis | File Difference Tracking | Automation of tracking the difference in different document versions. | 1 | 2 |
| | Error Detection | Detecting potential issues or errors within documents. | 1 | 0 |
| | Document Classification | Categorizing legal documents into predefined classes based on their content and characteristics. | 10 | 0 |
| | Document Management | Effective organization, storage, and collaboration of legal documents. | 1 | 4 |
| Document Development | Contract Generation | Automatic generation of all kinds of legal contracts. | 1 | 10 |
| | Enrichment of Documents | Adding additional information, annotations, or references to enhance legal documents. | 2 | 1 |
| | Summarization | Producing a shorter version of a lengthy legal text. | 5 | 3 |
| Information Processing | Anonymization | Removing or masking sensitive data and PII from documents. | 1 | 2 |
| | Information Extraction | Extracting important information from documents such as keywords. | 12 | 8 |
| | Document Retrieval | Search and retrieval of documents from knowledge bases. | 1 | 0 |
| Legal Dispute Resolution | Legal Decision Making | Automating the algorithmic decision-making process. | 1 | 7 |
| | Legal Reasoning | Assisting with argumentation and logical reasoning based on legal rules. | 5 | 2 |
| | Recommendation from Previous Court Rulings | Helping legal professionals develop effective strategies by analyzing insights from previous legal decisions. | 2 | 2 |
| Legal Assistance | Digital Assistant | Conversational agents that can help lawyers in their work. | 1 | 5 |
| | Question Answering | Answering legal questions based on given context or knowledge bases. | 3 | 1 |
| | Ranking of Lawyers | Automatic ranking of lawyers based on experience, success, expertise. | 1 | 0 |
| Knowledge Management | Changes in Law | Tracking changes in law to keep up to date with the current jurisdiction. | 1 | 1 |
| | Database for Court Decisions | A comprehensive database of all past court rulings, verdicts, and decisions. Ideally publicly available. | 1 | 1 |
| | Law Firm Management Software | The main software within a law firm, used for case management billing, timekeeping, collaborating. | 1 | 12 |
| | **Total:** | **117** | **55** | **62** |

**Table 3. Categories of legal use cases discovered in the SLR and SSIs**

Other than grouping the papers discovered in the SLR by their NLP technology, we additionally identified specific legal use cases in each of the papers. This means connecting the practical NLP solution described in the paper with the actual work of legal practitioners, i.e., how the solution can benefit their daily tasks.

Legal use cases detected in the SLR were later complemented and expanded with additional legal use cases originating from SSIs. The interviewees were asked if they are already using some NLP tools to help their work and which types of tasks they would like to see automated or assisted with in their work. After collecting all the legal use cases, we grouped them into seven overarching categories. This was done by following the literature and knowledge of legal experts collaborating on the project. Table 3 shows the categories, specific legal use cases with descriptions, and the number of mentions in the SLR and the SSIs.

The most popular shared use case is information extraction, which encompasses identifying entities and insights from long legal text. For most use cases, the number of mentions in the literature closely mirrored that from the interviews. Still, there are some use cases where there are discrepancies. Contract generation is an example of a use case found only once in our SLR, while it was mentioned by 10 different interviewees as a use case they would like to see automated in their daily work. On the other hand, document classification was found ten times in the literature, but not brought up at all in the interviews. This is also due to the fact that classification is often just one component of a larger use case such as search or topic discovery. While it is researched in NLP, its benefits are not so obvious for legal practitioners.

## *ELSA Concerns*

Both the SLR and interviews brought to light a set of ELSA concerns associated with the adoption of Legal NLP tools. Only around one-third of discovered papers (18 out of 49) mentioned ELSA in some form. In these 18 papers, ELSA typically came in a final appendix section on the limitations and ethical risks of deploying the described NLP tools. An example statement is "*In the triage process, there are risks associated with naive use of model predictions, including fairness across different user demographics*" (Mistica et al. 2021, p. 217). A total of 8 different aspects were mentioned in those 18 papers.

When looking at the interviews, ELSA concerns were mentioned by all interviewees. We identified 79 mentions between the SLR and SSIs. Among the interviewees, 10 said they have considerable ELSA concerns, while 8 said they have little to no ELSA concerns. Tables 4-6 show the identified ELSA concerns of using legal NLP solutions in legal work, with the total count mentioned for both the SLR and SSIs.

### Ethical aspects

| Ethical Aspect | Description | SSI | SLR |
|---|---|---|---|
| Empower Support, Not Final Verdict | Legal Tech solutions should only be used to support legal professionals in their decision-making process, while humans should be the ones making final decisions and actions. | 10 | 3 |
| Artificially Created Opinion | Models trained on data from a different country could generalize and create opinions that are not in line with the values of the nation and laws where they are actually used. | 6 | 0 |
| Threat of Discrimination | Due to existing biases in training data, it is possible that not everyone is treated equally in their access to justice. | 5 | 3 |
| Black-Box Principle | The inner workings and decisions of models are often not understandable to humans. | 4 | 1 |
| Transparency | The data used for training models and the source code are not always public. | 3 | 3 |
| Technology Misuse | Any technology has a potential to be misused for malicious or unethical purposes. | 1 | 0 |
| Monopoly | Single dominant entity having control over all the relevant Legal Tech or NLP solutions and making all the decision related to it. | 1 | 0 |
| | **Total:** | **33** | **10** |

**Table 4. Ethical aspects of using Legal NLP as mentioned in the SSIs and SLR**

Ethical aspects include human values like morality, fairness, equity, privacy, and transparency. Concretely, the most common aspect is focused on the fact that Legal Tech should serve as an assistive tool for humans and not be making any final decisions. As I-18 said: "*It is very important that the actual legal decision-making is done by a human being.*" Another finding is the fear of artificially created opinions, where trained AI models would reproduce their training data and thus "*not represent the values of the people and the nation*" (I-4). Similarly, I-9 argues: "*Law is inherently unstable. It is always up to the zeitgeist of the people, of the judges*". This shows the need for better alignment of technology with human values and laws.

Other ethical concerns center around the threat of discrimination and unequal treatment due to biases in training data and the misuse of technology for malicious purposes. The lack of interpretability of the inner workings of models is another common concern ("*It would be unethical to provide a tool to judges that they do not understand and make them rely on it*", I-6), together with a lack of transparency about the source code of the tools. Finally, the threat of monopolization was expressed by a single interviewee.

**Legal aspects**

| Legal Aspect | Description | SSI | SLR |
|---|---|---|---|
| Legal Compliance | The technical solutions used should follow all the mandatory laws and legal requirements. | 11 | 1 |
| Data Protection | Protecting the data of clients in legal firms is of high importance in their daily work. | 5 | 3 |
| Faulty Results | Models producing wrong results could lead to legal problems and the professionals using them would be responsible for it. | 4 | 0 |
| Compliant Creation | Technology should be created to be legally compliant from the start of its lifecycle. | 2 | 0 |
| | **Total:** | **22** | **4** |

**Table 5. Legal aspects of using Legal NLP as mentioned in the SSIs and SLR**

Legal aspects are concerned with regulatory compliance, liability, data protection, and intellectual property. Legal compliance was a very common concern, and many interviewees seemingly valued it more than other aspects such as security or transparency ("*Above all, it is necessary to be legally compliant, even if it is not completely secure*", I-7). This is tied to compliant creation, which puts emphasis on creating solutions that are legally compliant, a point that is important to consider in the design of Legal Tech going forward.

Client data protection is another important concern in law firms ("*The use of cloud is effectively prohibited for us, because no data should leave the premises. The same would apply for legal AI tools*", I-11). Fear of incorrect results was also mentioned, including the liability for errors that AI models would make. I-9 explains it: "*We are very sensitive when it comes to errors, since we would be personally liable for any mistakes models make. For me, an accuracy of over 98% is needed to call it precise enough.*"

**Social aspects**

| Social Aspect | Description | SSI | SLR |
|---|---|---|---|
| Transformation of Work | The pace of development of new AI solutions could lead to a change in the structure of the legal workforce and skills needed in legal jobs. | 10 | 9 |
| Empathy | Law inherently works with humans and having empathy and understanding for legal clients is an important aspect that machines usually do not possess. | 7 | 0 |
| Access to Justice | A positive aspect of using Legal Tech would be making law more accessible to citizens. | 3 | 7 |
| Readiness of Society | If society is not ready to adopt AI, there will be a pushback to their use in legal work. | 2 | 0 |
| Dependability on Technology | Relying too much on technology could lead to a loss of skills of legal professionals and to problems when the technology is not available (e.g., servers are down). | 1 | 0 |
| Peer Pressure | To be competitive, smaller firms could be forced to use tools for which they are not ready. | 1 | 0 |
| | **Total:** | **24** | **16** |

**Table 6. Social aspects of using Legal Tech as mentioned in the SSIs and SLR**

Social aspects concentrate on factors like inclusivity, security, and public perception. The most common aspect is the transformation of work that will follow with the greater adoption of AI in law. I-3 sees the role of lawyers changing to the role of a reviewer, where in the future "*clients will hand in the documents generated by legal AI tools and we just take responsibility for it*". When it comes to existential threats, I-14 comments: "*We have to be concerned a bit about the future and our own economic existence*", while I-13 is of the opposite opinion: "*Machines will never replace humans. I don't feel threatened in any way regarding my existence*". Other fears include becoming too dependent on new technology and peer pressure to adopt the new tools despite unpreparedness, in order to stay relevant in the legal landscape.

A commonly mentioned aspect was empathy, which included all concerns related to the lack of human touch and personalization of legal technology. The problem of trust was brought up by I-11, who shared: "*I had clients open up to me about money in hidden Swiss bank accounts or about children from extramarital affairs. Would they admit something personal like this to a machine?*" One more aspect is making legal services more accessible to the public through technology. Finally, societal acceptance of new technology is needed for its wider adoption. As I-2 concludes, "*There is great potential in the automation of the law enforcement. But as long as the society doesn't want it, it simply won't happen.*"

## Discussion

An interesting finding comes with the mapping of individual NLP categories to the identified use cases. In many use cases, such as Legal Reasoning, the successful adoption of such a Legal Tech solution would necessitate a confluence of NLP technologies, making direct mapping quite difficult. Another example comes with anonymization, which falls under the technical category of Information Processing, although one may argue that it pertains to Trustworthiness as well. Investigating the complex relationship between technologies and use cases should be further investigated not only to refine the catalog of use cases, but also to boost practical applicability and raise awareness of the potential uses of a given NLP technology. Thus, we see it as an important task to connect the identified NLP categories with their associated use cases.

Another interesting finding comes with the numerous ELSA concerns expressed by legal experts, as well as the literature. In each of our three aspect categories, at least one aspect was mentioned by a majority of the interviewees, and 11 of the 15 concerns overall were mentioned more than once. This highlights that legal practitioners often share similar concerns with the adoption of Legal Tech. Looking at the most commonly shared concerns, practitioners agree that while Legal Tech shows promise, it is ultimately a support tool, rather than a completely disruptive movement to replace legal roles. Nevertheless, the social implications of Legal Tech are clear, and its potential to transform the legal workplace was indeed reiterated. An interesting finding also comes with the legal concerns expressed by the expert interviewees, as three of the four concerns revolve around compliance and data protection, an important indicator of the strong enforcement presence of modern-day regulations such as the GDPR. Additionally, one can observe that the identified literature makes no mention at all of multiple concerns raised across the three categories.

Many of the other concerns expressed by the legal experts echo general concerns about the proliferation of advanced NLP technologies. In particular, questions of transparency, bias, privacy, and misuse are very much aligned with current research on the potential dangers of ubiquitous AI. While these certainly should not serve as major deterrents to the continued development of NLP-driven legal technologies, these ELSA concerns expressed make salient the need for the responsible and socially conscious design of Legal Tech.

An important insight comes to light with a closer analysis of the specific use cases mentioned by many of the expert interviewees. While some use cases, such as Digital Assistants, rely on the use of large language models (LLMs) and other "in demand" NLP technologies, many pertain to the much more day-to-day tasks of legal professionals, such as information processing and knowledge management. The divide between state-of-the-art NLP research and the needs of legal practitioners thus raises the question of whether the objectives of Legal NLP research and practical Legal Tech use cases are aligned. In particular, it becomes important to find the balance between advancing the state-of-the-art and maintaining practical usability.

Following from the above, we identify a significant gap between the current trend of NLP technologies that focuses on the development and optimization of LLMs, the need for such tools as expressed by our interviewees is markedly low. This gap invites future research to survey the current state of adoption of LLMs in legal practice; moreover, identifying the major obstacles to such adoption would be highly valuable.

Another remaining gap pertains to the relationship between the identified use cases and the ELSA concerns. While the numerous ELSA concerns brought up by the interviewees serve as an interesting point of analysis on their own, a more meaningful discussion can follow from deeper insight into which use cases evoke which ELSA concerns. With such insight, researchers and practitioners alike could be more aligned on the critical points to address in order to advance Legal Tech while minimizing its perceived riskiness.

## Conclusion

In this work, we investigate the use cases of NLP in Legal Tech, grounded in a systematic investigation of technical solutions as provided in the literature. With these technologies as a baseline, we conducted a series of 18 semi-structured interviews, which brought to light 22 use cases for Legal Tech. In particular, we focused on the ethical, legal, and social concerns mentioned by the interviewees, to highlight the potential barriers hindering the wider adoption of Legal Tech solutions.

The limitations of the presented work mainly originate from the targeted focus of our interview study: the legal field in Germany. Further work to generalize our findings to a broader geographical region would be well-served to validate our findings. In addition, the relatively small number of interviewees should be broadened to include a diverse range of legal professions. Nevertheless, the observed range of roles in our interview study serves as a starting point, with at least five distinct legal roles. Finally, the threat to internal validity in terms of researcher bias was mitigated by employing a diverse team of researchers, all of whom contributed to the analysis of selected literature, as well as the review and coding of interview transcripts.

With the fast-paced proliferation of available NLP technologies, the practical implications of our work become clear, as without clearly defined use cases, the transition of Legal Tech from research to practice may be hindered. Furthermore, the identification of relevant ethical, legal, and social aspects in the design and adoption of Legal Tech solutions is crucial to both researchers and practitioner, in order to understand the implications that developing and implementing such technologies carries. This is important in bringing the innovative research in the NLP field into the hands of legal professionals.